\documentclass[12pt]{article}

\usepackage[utf8]{inputenc}
\usepackage[T1]{fontenc}
\usepackage{amsmath,amssymb}
\usepackage{graphicx}
\usepackage{hyperref}
\usepackage{url}
\usepackage{booktabs}
\usepackage{tabularx}
\usepackage{natbib}
\usepackage[margin=1in]{geometry}
\usepackage{setspace}
\usepackage{authblk}

\hypersetup{
    colorlinks=true,
    linkcolor=blue,
    citecolor=blue,
    urlcolor=blue
}

\title{Counterweights and Complementarities: The Convergence of AI and Blockchain Powering a Decentralized Future}

\author[1]{Yibai Li \thanks{yibai.li@scranton.edu}}
\author[2]{Zhiye Jin \thanks{zjin@m.marywood.edu}}
\author[1]{Xiaobing (Emily) Li}
\author[3]{K.~D.~Joshi}
\author[4]{Xuefei (Nancy) Deng}

\affil[1]{University of Scranton}
\affil[2]{Marywood University}
\affil[3]{University of Nevada, Reno}
\affil[4]{California State University, Dominguez Hills}

\date{}

\begin{document}

\maketitle

\begin{abstract}
This editorial addresses the critical intersection of artificial intelligence (AI) and blockchain technologies, highlighting their contrasting tendencies toward centralization and decentralization, respectively. While AI, particularly with the rise of large language models (LLMs), exhibits a strong centralizing force due to data and resource monopolization by large corporations, blockchain offers a counterbalancing mechanism through its inherent decentralization, transparency, and security. The editorial argues that these technologies are not mutually exclusive but possess complementary strengths. Blockchain can mitigate AI's centralizing risks by enabling decentralized data management, computation, and governance, promoting greater inclusivity, transparency, and user privacy. Conversely, AI can enhance blockchain's efficiency and security through automated smart contract management, content curation, and threat detection. The core argument calls for the development of ``decentralized intelligence'' (DI)---an interdisciplinary research area focused on creating intelligent systems that function without centralized control.
\end{abstract}


\section{Introduction}

Artificial intelligence (AI) and blockchain technologies represent two of the most transformative forces of the 21st century. AI technology enables computers and machines to simulate human learning, comprehension, problem-solving, and decision-making, while blockchain is a decentralized, secure, and transparent ledger technology for recording transactions and data. Often considered distinct technologies, AI and blockchain can develop synergy, presenting both challenges and opportunities. The rapid advancement of AI has ushered in an era of unprecedented technological capabilities, impacting various sectors from healthcare to finance. However, this progress has also raised concerns about the centralization of power, data monopolization, and ethical implications \citep{bender2021dangers, bommasani2021opportunities}. Concurrently, blockchain technology has emerged as a disruptive force, promising decentralization, transparency, and enhanced security \citep{asif2024leveraging}. This editorial explores the inherent tensions and synergistic potential between AI and blockchain, advocating for a decentralized approach to AI research and development. We argue that blockchain, with its decentralized nature, can serve as a critical counterweight to the centralizing tendency of AI. This editorial calls for research into ``decentralized intelligence'' (DI)---an interdisciplinary research area focused on creating intelligent systems that function without centralized control. The editorial concludes with proposing potential research questions exploring how blockchain can address AI issues of data and resource monopolization, as well as the concentration of power and control.

\section{AI and Blockchain as Counterweights}

AI and blockchain, while both revolutionary technologies, often exhibit opposing tendencies. This section delves into the fundamental differences that position them as counterweights in the evolving technological landscape.

\subsection{AI's Centralizing Tendency vs.\ Blockchain's Decentralizing Nature}

Perhaps the most fundamental difference between AI and blockchain lies in their inherent different tendencies toward centralization. AI tends toward centralization, but blockchain is built on the principle of decentralization. This difference is so stark that some observers have framed AI and blockchain in starkly political terms. As Ali Yahya \citep{a16z2023} put it, ``AI is communist,'' and blockchain, such as crypto, ``is libertarian.''

The centralization problem of AI refers to the concentration of development, access, and control of these powerful AI systems in the hands of a few corporations, research institutions, and governments. This centralization raises concerns about accessibility, transparency, ethics, and economic influence \citep{bender2021dangers, bommasani2021opportunities}.

AI, by its nature, tends toward centralization because AI models often require massive amounts of data and computational power, which are typically controlled by large corporations \citep{wang2024aiarena}. AI thrives on data: the more data an entity has, the better its AI models perform, leading to a concentration of power. The development and deployment of AI systems are often driven by large organizations, resulting in centralized control \citep{zyskind2024secure}. Moreover, AI is seen as a sustaining innovation that strengthens the existing business models of tech giants. The centralizing tendency of AI has given rise to three problems.

\subsubsection{Data Monopolization}

Large language models (LLMs) rely on massive amounts of data for effective training and operation, but this data is increasingly concentrated in the hands of a few major technology companies. This data monopolization has several detrimental consequences. It creates significant barriers for researchers, developers, and smaller firms outside these tech giants, limiting their access to the high-quality, large-scale datasets needed to train competitive models. This lack of data access hinders innovation and can lead to unfair competition in the AI industry \citep{weidinger2021ethical}. Furthermore, selective data inclusion by these dominant companies can introduce biases into the models, influencing their outputs and potentially perpetuating existing societal inequalities.

\subsubsection{Resource Monopolization}

Beyond data, the development and deployment of LLMs require immense computational resources, leading to resource monopolization. For example, the training of state-of-the-art models like GPT-4 was reported to cost more than \$100 million \citep{openai2023gpt4}. AI model training necessitates access to high-performance GPUs, vast data storage capabilities, and substantial electricity consumption. The ongoing costs of maintaining and fine-tuning these models further exacerbate the financial burden. These high costs, combined with the required, specialized infrastructure, effectively restrict the development of cutting-edge LLMs to a small number of exceptionally wealthy corporations, creating a ``winner-takes-all'' dynamic in the market \citep{floridi2020gpt3}.

\subsubsection{Concentration of Power and Control}

As a direct consequence of both data and resource monopolization, the development, deployment, and control of LLMs are becoming increasingly concentrated in the hands of a few powerful tech companies, such as OpenAI, Google, Meta, and Microsoft. This concentration of power and control presents several significant challenges. Smaller businesses and independent researchers are often excluded from accessing and utilizing cutting-edge AI, stifling innovation and competition. Furthermore, this concentration gives these few companies significant influence over policy-making and regulatory frameworks related to AI, potentially shaping the future of the technology to their advantage \citep{solaiman2021process}. Finally, users become increasingly dependent on a handful of corporations for AI-driven services, raising concerns about autonomy and potential misuse of power.

In contrast, blockchain is fundamentally about decentralization. Designed to enable cooperation among individuals without the need for intermediaries or centralized authorities, blockchain systems aim to distribute control and power across a network of users rather than concentrating it in a single entity \citep{bassey2024peer}. Moreover, blockchain seeks to give individuals more control over their data, assets, and digital identities \citep{asif2024leveraging}.

\subsection{AI Reinforces Incumbents, but Blockchain Disrupts Incumbents}

AI is seen as a sustaining innovation, which means it primarily improves existing technologies and business models. This tendency allows large tech companies, which already possess the necessary resources like data and computational power, to further strengthen their market position and dominance \citep{petrosino2025zero}. The benefits of AI innovations are often captured by the companies that control the platforms and data, further entrenching their dominant position \citep{din2024building}.

Blockchain is characterized as a disruptive innovation, which means it challenges existing power structures and business models \citep{liu2024application}. Blockchain's drive toward decentralization and individual empowerment directly opposes the centralized models of many existing tech companies. Described as being driven by ``rebels'' and ``fringes,'' blockchain challenges established norms and powers and aims to create a fairer system where power is more distributed \citep{lustenberger2024dao}.

\subsection{AI Erodes Privacy vs.\ Blockchain Enhances Privacy}

AI's functionality relies heavily on data collection and analysis, creating a strong incentive to gather as much user information as possible. This data-centric approach can lead to scenarios where individual privacy is compromised, moving toward a ``panopticon'' scenario, where individuals are constantly monitored \citep{johnson2014watching}. The hunger for data by AI systems pushes toward less individual privacy.

Blockchain, on the other hand, emphasizes user sovereignty and control over data, aiming to move toward increased privacy \citep{awad2025protecting}. The focus is on enabling users to own and control their data within AI systems, pushing back against data aggregation. Blockchain's focus on decentralization and individual empowerment is directly aligned with the goal of increasing privacy, giving individuals more control over their digital lives. Blockchain moves in the opposite direction of AI in terms of privacy by providing tools for users to control their data.

\subsection{AI Creates Infinite Media Abundance vs.\ Blockchain Defends Human Value}

AI's generative capabilities result in a vast amount of easily produced content. The sheer volume of AI-generated media can be overwhelming, making it difficult to distinguish between human-created content and AI-generated content \citep{divya2024transforming}. The ease of AI content creation can lead to a flood of media, making it challenging to assess value.

Blockchain provides mechanisms to help distinguish human-created content and maintain its value \citep{picha2024blockchain}. NFTs (non-fungible tokens) and provenance tracking can be used to establish the authenticity and origin of digital content, preserving its value \citep{hasan2024nfts}. Blockchain helps to identify and preserve what is genuinely human in a world saturated by AI-generated media. By providing ways to verify authenticity, blockchain aims to ensure that human creations are not inundated with the endless stream of AI-produced media.

\section{AI and Blockchain Complementarities and Mutual Benefits}

Despite the inherent tensions between AI's centralizing force and blockchain's decentralizing ethos, these two technologies possess complementary strengths that, when combined, can mitigate their weaknesses. Blockchain, with its decentralized nature, offers a potential solution to many of the challenges posed by AI with its centralizing tendency. By facilitating decentralization, blockchain can distribute AI model training and inference across a network, mitigating the risks of centralized control and data monopolies \citep{asif2024leveraging}. Furthermore, blockchain's immutable ledger enhances trust and transparency in AI systems by enabling verifiable audits of data inputs and model outputs, thereby promoting accountability.

A significant development in this area is zero-knowledge machine learning (ZKML), which utilizes blockchain to verify the integrity of AI computations without revealing the underlying data. This innovation, as detailed by \citet{peng2025survey}, offers a robust solution for maintaining data confidentiality while enhancing trustworthiness. Blockchain also empowers users with greater data privacy, enabling secure, privacy-preserving data sharing for AI training and providing individuals with granular control over access to their personal information. Additionally, blockchain-based proof-of-humanity mechanisms enhance authenticity verification, combating AI-generated misinformation by confirming that content is genuinely human-generated.

Additionally, AI can significantly enhance the functionality and security of blockchain systems. AI-driven automation in smart contract code generation and vulnerability detection increases efficiency and security in blockchain applications \citep{barbara2024automatic, rahman2024riskaichain}. AI's role in content curation is particularly valuable for decentralized platforms, where it can effectively filter spam and misinformation, ensuring platform integrity and reliability.

AI also substantially benefits security within blockchain systems. The application of AI for transaction monitoring and miner extractable value (MEV) \citep{xue2022preventing} defense strengthens resistance against malicious actors, protecting blockchain users from exploitation. Moreover, AI can detect and flag deepfakes and other forms of misinformation, safeguarding blockchain-based platforms from harmful or false information \citep{lokhande2024artificial}.

These combined benefits illustrate a powerful, symbiotic relationship between AI and blockchain technologies. Each technology addresses the limitations of the other, fostering a mutually beneficial partnership that enhances overall effectiveness and security, as evidenced by recent studies \citep{barbara2024automatic, rahman2024riskaichain}.

\section{Decentralized Intelligence: The Definition and Call for Research}

Decentralized intelligence (DI) research is the interdisciplinary study of methodologies, algorithms, and architectures that enable intelligent systems to function without centralized control. The origins of decentralized intelligence can be traced back to the rise of distributed computing in the 1950s--1980s. Early digital computing introduced parallel processing and distributed systems, which enabled multiple processors to work on tasks simultaneously. In the 1980s--1990s, multi-agent systems (MAS) emerged, allowing independent computational agents to interact and collaboratively solve complex problems in a decentralized manner \citep{mueller2025distributed}. In the 1990s--2000s, the rise of internet technologies accelerated developments in decentralized intelligence, especially through swarm intelligence inspired by natural systems and peer-to-peer networks like Napster (1999) and BitTorrent (2001). Grid computing also emerged during this period, distributing computing tasks across multiple independent nodes. The introduction of blockchain technology with Bitcoin in 2009 established trustless decentralized systems, inspiring AI researchers to adopt blockchain principles for secure distributed intelligence. Google's Federated Learning in 2017 further advanced decentralized AI by enabling machine learning across devices without centralized data storage \citep{yang2019federated}.

From November 30, 2022, to the present, the explosive emergence of LLMs has exacerbated the issue of AI centralization, presenting both significant challenges and novel research opportunities within the field of DI.

We propose that decentralized intelligence demands a new ecosystem surrounding AI, an ecosystem that, for instance, consists of development platforms, research consortia, regulatory frameworks, and standardization bodies. We call for the following research initiatives:

\begin{enumerate}
    \item \textbf{Government-Funded Open-AI Systems:} Governments should invest in the development of open-source AI models and datasets to provide a public alternative to proprietary AI systems controlled by corporations.
    \item \textbf{Research Consortia:} Universities, research institutions, and industry partners should collaborate in consortiums to conduct research on decentralized AI, sharing resources and expertise.
    \item \textbf{Regulatory Frameworks:} Regulatory frameworks should be developed to address the unique challenges of decentralized AI, including data privacy, algorithmic accountability, and liability. These regulations should be designed to promote innovation while mitigating risks.
    \item \textbf{Decentralized Data Cooperatives:} Encourage the formation of data cooperatives where individuals can pool their data and collectively control its use for AI training, ensuring fair compensation and data sovereignty.
    \item \textbf{Standardization Bodies:} International bodies tasked with creating interoperability standards and best practices for decentralized AI systems, ensuring compatibility and preventing vendor lock-in.
    \item \textbf{Open-Source Development Platforms:} Platforms like a decentralized GitHub for AI, where researchers can collaboratively develop, test, and deploy decentralized AI models and applications.
\end{enumerate}

We also advocate for continued research into ``decentralized intelligence,'' particularly through leveraging blockchain technology. Potential research topics include, but are not limited to, the areas described in Table~\ref{tab:research_questions}.

\begin{table}[htbp]
\centering
\caption{Potential Research Questions on Blockchain-powered Decentralized AI}
\label{tab:research_questions}
\small
\begin{tabularx}{\textwidth}{>{\raggedright\arraybackslash}p{0.42\textwidth} X}
\toprule
\textbf{Research Question} & \textbf{Potential Implications and Mechanisms} \\
\midrule
How can blockchain-enabled decentralized AI marketplaces address the concentration of power in AI? & Decentralized marketplaces could empower smaller developers, fostering innovation and diversity by bypassing corporate gatekeepers. Blockchain may also enhance federated learning by securing updates and ensuring data provenance, enabling distributed training \citep{bensasson2014zerocash, mcmahan2017communication}. \\
\addlinespace
How can blockchain mitigate data monopolization? & Blockchain-based data cooperatives and DAOs could allow individuals to collectively own and manage data, challenging large corporations and promoting fairer data-sharing \citep{hassan2021decentralized}. Data marketplaces on the blockchain could facilitate direct data sharing/sales with AI developers \citep{wirth2018privacy}. \\
\addlinespace
What is blockchain's potential to lower economic and competitive barriers to AI innovation? & Tokenized access to computing resources, facilitated by blockchain, could democratize AI model training/deployment. Micropayments via blockchain could introduce new business models, lowering entry barriers \citep{swan2015blockchain}. \\
\addlinespace
How can blockchain mitigate AI-driven privacy erosion? & Blockchain-based decentralized identity (DID) systems and verifiable credentials could give individuals more control over their data \citep{tobin2017inevitable}. Combining blockchain with zero-knowledge proofs (ZKPs) can further enhance privacy in AI data verification \citep{bensasson2014zerocash}. \\
\addlinespace
How might NFTs help establish provenance and authenticity for human-created content in the age of AI-generated media? & NFTs could establish provenance and authenticity for human-created content, preserving its value. Decentralized, blockchain-based platforms could directly connect creators and audiences, reshaping the creative economy \citep{wang2021nft, atzori2015blockchain}. \\
\bottomrule
\end{tabularx}
\end{table}

\section{Conclusion}

AI centralization is a socio-technical problem. The mere use of a blockchain does not guarantee decentralization; instead, it depends heavily on its design, governance, and implementation \citep{yaga2018blockchain}.

The pursuit of decentralized intelligence represents a significant shift in the AI landscape, aiming to address the inherent risks and limitations of centralized systems. By leveraging innovations like blockchain, we can create AI ecosystems that are powerful, efficient, equitable, transparent, and aligned with broader societal values. This requires a concerted, interdisciplinary effort, encompassing technological development, ethical considerations, and robust governance structures.

We encourage IS scholars to contribute to the pursuit of decentralized intelligence by engaging and supporting the dedicated mini-track on Responsible Approaches to Blockchain, Cryptocurrency, and Fintech in the IT, Social Justice, and Marginalized Contexts track at the Hawaii International Conference on System Sciences 2024--2026, which would allow the IS community to attract, promote, and grow this research systematically and deliberately \citep{joshi2024it}.

\bibliographystyle{apalike}

\end{document}